\pdfoutput=1

\documentclass[11pt]{article}
\usepackage{float}
\usepackage{acl2021}

\usepackage{times}
\usepackage{latexsym}

\usepackage[T1]{fontenc}

\usepackage[utf8]{inputenc}

\usepackage{microtype}
\usepackage{amsmath, amssymb}
\usepackage{booktabs}
\usepackage{graphicx}
\usepackage{multirow}
\usepackage[noend]{algpseudocode}
\usepackage{amsmath}
\usepackage{algorithmicx,algorithm}
\title{ATP: AMRize Then Parse! Enhancing AMR Parsing with PseudoAMRs}

\author{
 Liang Chen$^1$$\footnotemark[1]$, \ Peiyi Wang$^1$$\footnotemark[1]$,\  Runxin Xu$^1$,\  Tianyu Liu$^2$, \\ \textbf{Zhifang Sui$^1$,\  Baobao Chang$^1$$\footnotemark[2]$}
\\ 
$^1$ Key Laboratory of Computational Linguistics, Peking University, MOE, China \\
$^2$ Tencent Cloud Xiaowei \\
 \texttt{leo.liang.chen@outlook.com; wangpeiyi9979@gmail.com} \\
 \texttt{runxinxu@gmail.com; rogertyliu@tencent.com} \\
 \texttt{szf,chbb@pku.edu.cn}
}

\date{}

\begin{document}
\newcommand{\tianyu}[1]{\textcolor{green}{\bf \small [ #1 --tianyu]}}
\newcommand{\cl}[1]{\textcolor{orange}{\bf \small [ #1 --Chen Liang]}}
\newcommand{\py}[1]{\textcolor{blue}{\bf \small [ #1 --Peiyi]}}

\maketitle
\begin{abstract}

As Abstract Meaning Representation (AMR) implicitly involves compound semantic annotations, we hypothesize auxiliary tasks which are semantically or formally related can better enhance AMR parsing. We find that 1) Semantic role labeling (SRL) and dependency parsing (DP), would bring more performance gain than other tasks e.g. MT and summarization in the text-to-AMR transition even with much less data. 2) To make a better fit for AMR, data from auxiliary tasks should be properly ``AMRized'' to PseudoAMR before training. Knowledge from shallow level parsing tasks can be better transferred to AMR Parsing with structure transform. 3) Intermediate-task learning is a better paradigm to introduce auxiliary tasks to AMR parsing, compared to multitask learning. From an empirical perspective, we propose a principled method to involve auxiliary tasks to boost AMR parsing. Extensive experiments show that our method achieves new state-of-the-art performance on different benchmarks especially in topology-related scores. Code and models are released at \url{https://github.com/PKUnlp-icler/ATP}.

\end{abstract}
\renewcommand{\thefootnote}{\fnsymbol{footnote}}
\footnotetext[1]{Equal Contribution.}
\footnotetext[2]{Corresponding Author.}
\section{Introduction}
\begin{figure}[t]
    \centering
    \includegraphics[width=1\linewidth]{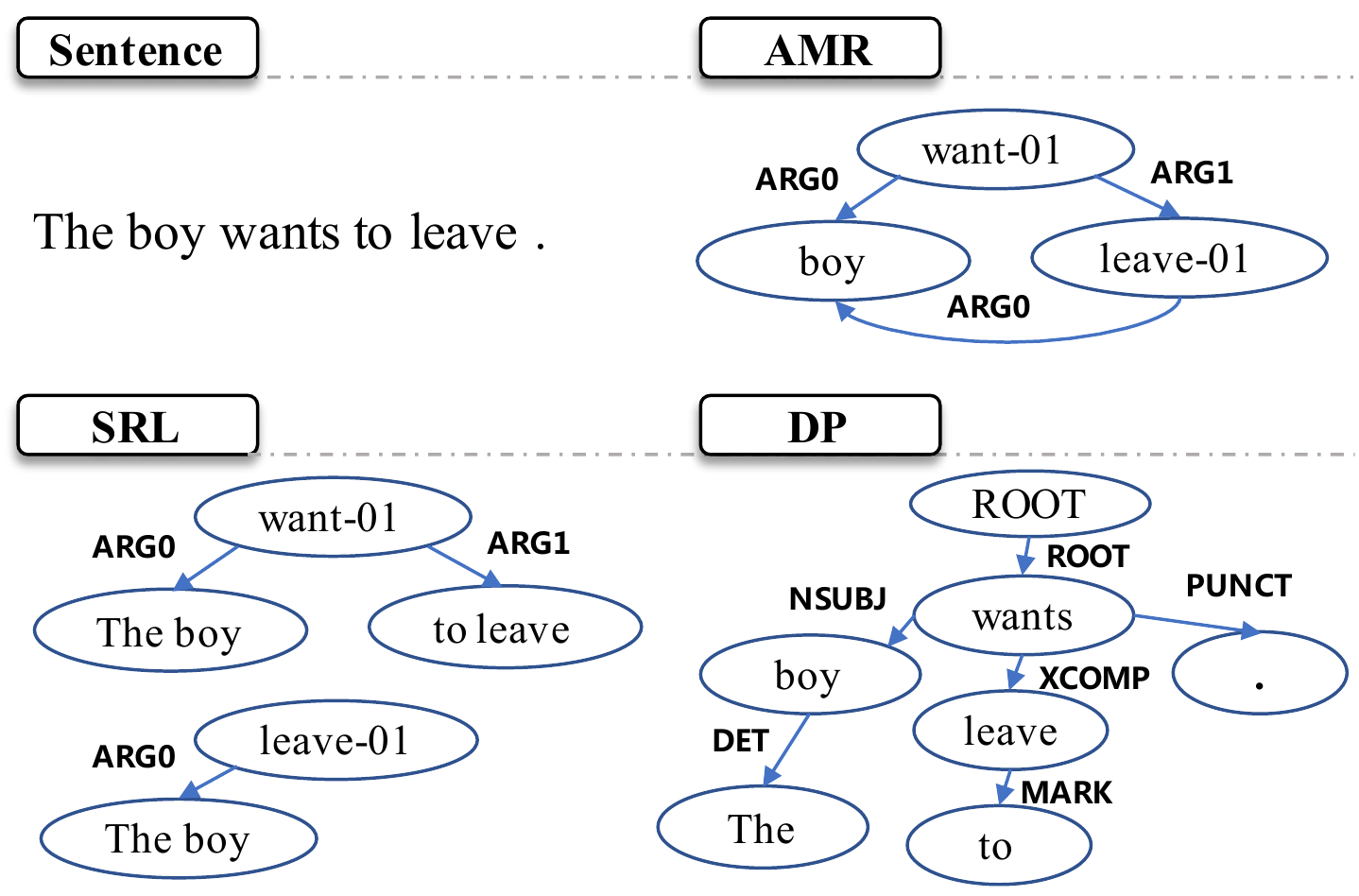}
    \caption{The Abstract Meaning Representation (AMR), Semantic Role Labeling (SRL), and Dependency Parsing (DP)  structure of the sentence ``The boy wants to leave.''}
    \label{fig:srl_dp_amr}
\end{figure}

Abstract Meaning Representation (AMR) \citep{ban-AMR} parsing aims to translate a sentence to a directed acyclic graph, which represents the relations among abstract concepts as shown in Figure~\ref{fig:srl_dp_amr}. AMR can be applied to many downstream tasks, such as information extraction \cite{rao-amr-ie, wang-amr-ie, zhang-amr-ie}, text summarization,  \cite{liao-amr-tm, hardy-amr-tm} question answering \cite{mitra-amr-qa, sacha-amr-qa} and dialogue modeling \citep{Bonial2020DialogueAMRAM}.

Recently, AMR Parsing with the sequence-to-sequence framework achieves most promising results \citep{xu-seqpretrain,bevil-spring}. Comparing with transition-based or graph-based methods, sequence-to-sequence models do not require tedious data processing and is naturally compatible with auxiliary tasks \citep{xu-seqpretrain} and powerful pretrained encoder-decoder models \citep{bevil-spring}. Previous work \citep{xu-seqpretrain,Wu2021ImprovingAP} has shown that the performance of AMR parser can be effectively boosted through co-training with certain auxiliary tasks, e.g. Machine Translation or Dependency Parsing. 

However, when introducing auxiliary tasks to enhance AMR parsing, we argue that three important issues still remain under-explored in the previous work. \textbf{1) How to choose auxiliary task?} The task selection is important since loosely related tasks may even impede the AMR parsing according to \citet{Damonte2021OneSP}. However, in literature there are no principles or consensus on how to choose the proper auxiliary tasks for AMR parsing. Though previous work achieves noticeable performance gain through multi-task learning, they do not provide explainable insights on why certain task outperforms others or in which aspects the auxiliary tasks benefit the AMR parser.
\textbf{2) How to bridge the gap between tasks ?} The gaps between AMR parsing and auxiliary tasks are non-negligible. For example, Machine Translation generates text sequence while Dependency Parsing (DP) and Semantic Role Labeling (SRL) produces dependency trees and semantic role forests respectively as shown in Figure~\ref{fig:srl_dp_amr}. Prior studies \citep{xu-seqpretrain,Wu2021ImprovingAP,Damonte2021OneSP} do not attach particular importance to the gap, which might lead the auxiliary tasks to \textit{outlier-task} \citep{multi-sy,cai_will_2017} in the Multitask Learning, deteriorating the performance of AMR parsing. 
\textbf{3) How to introduce auxiliary tasks more effectively?} After investigating different training paradigms to combine the auxiliary task training with the major objective (AMR parsing), we figure out that, although all baseline models     \citep{xu-seqpretrain,Wu2021ImprovingAP,Damonte2021OneSP} choose to jointly train the auxiliary tasks and AMR parsing with Multi-task Learning (MTL), Intermediate-task Learning (ITL) is a more effective way to introduce the auxiliary tasks for pretrained models. Our observation is also consistent with \citep{kun2020intermediate,poth2021intermediate}, which improve other NLP tasks with enhanced pretrained models.

\begin{figure}[!t]
    \centering
    \includegraphics[width=1\linewidth]{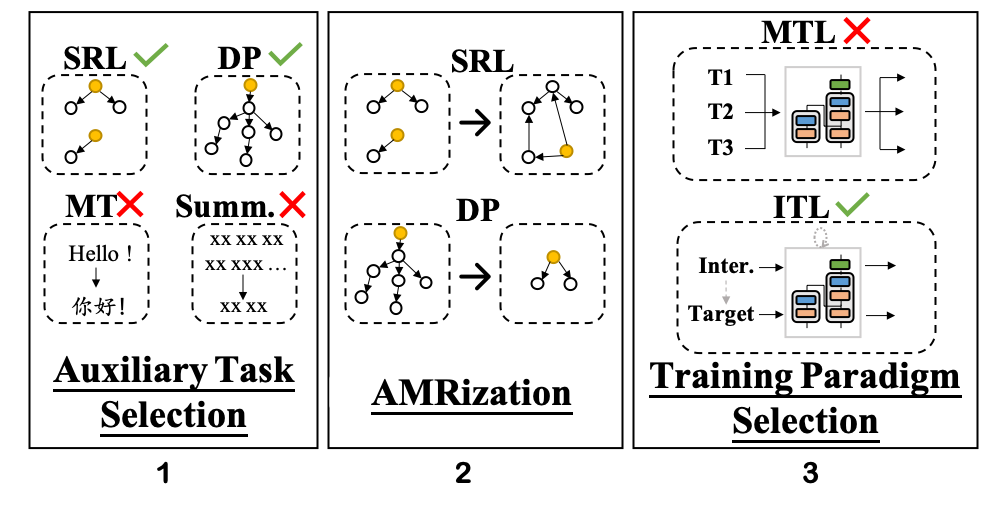}
    \caption{Illustration of methodology in this paper. We proposed a principled method to select, transform and train the auxiliary tasks.}
    \label{fig:procedure}
\end{figure}

In response to the above three issues, we summarize a principled method to select, transform and train the auxiliary tasks (Figure~\ref{fig:procedure}) to enhance AMR parsing from extensive experiments.
\textbf{1) Auxiliary Task Selection}. We choose auxiliary tasks by estimating their similarities with AMR from the semantics and formality perspectives. AMR is recognized as a deep semantic parsing task which encompasses multiple semantic annotations, e.g. semantic roles, name entities and co-references. As a direct semantic-level sub-task of AMR parsing, we select SRL as one auxiliary task. Traditionally, formal semantics views syntactic parsing a precursor to semantic parsing, leading to the mapping between syntactic and semantic relations. Hence we introduce dependency parsing, a syntactic parsing task as another auxiliary task. 
\textbf{2) AMRization.} Despite being highly related, the output formats of SRL, DP and AMR are distinct from each other.  
To this end, we introduce transformation rules to ``AMRize'' SRL and DP to PseudoAMR, intimating the feature of AMR. Specifically, through \textit{Reentrancy Restoration} we transform the structure of SRL to a graph and restore the reentrancy within arguments, which mimics AMR structure. Through \textit{Redundant Relation Removal} we conduct transformation in dependency trees and remove relations that are far from semantic relations in AMR graph.
\textbf{3) Training Paradigm Selection.} We find that ITL makes a better fit for AMR parsing than MTL since it allows model progressively transit to the target task instead of learning all tasks simultaneously, which benefits knowledge transfer \citep{multi-sy}. 

We summarize our contributions as follows: 
\begin{enumerate}
    \item Semantically or formally related tasks, e.g., SRL and DP, are better auxiliary tasks for AMR parsing compared with distantly related tasks, e.g. machine translation and machine reading comprehension.
    \item We propose task-specific rules to  AMRize the structured data to PseudoAMR. SRL and DP with properly transformed output format further improve AMR parsing. 
    \item ITL outperforms classic MTL methods when introducing auxiliary tasks to AMR Parsing. We show that ITL derives a steadier and better converging process during training.
\end{enumerate}

  Extensive experiments show that our method (PseudoAMR + ITL) achieves the new state-of-the-art of single model on in-distribution (85.2 Smatch score on AMR 2.0, 83.9 on AMR 3.0), out-of-distribution benchmarks. Specifically we observe that AMR parser gains larger improvement on the SRL(+3.3), Reentrancy(+3.1) and NER(+2.0) metrics\footnote{Computed on AMR 2.0 and 3.0 dataset.}, due to higher resemblance with the selected auxiliary tasks.  

\section{Methodology}
\begin{figure*}[t]
    \centering
    \includegraphics[width=1\linewidth]{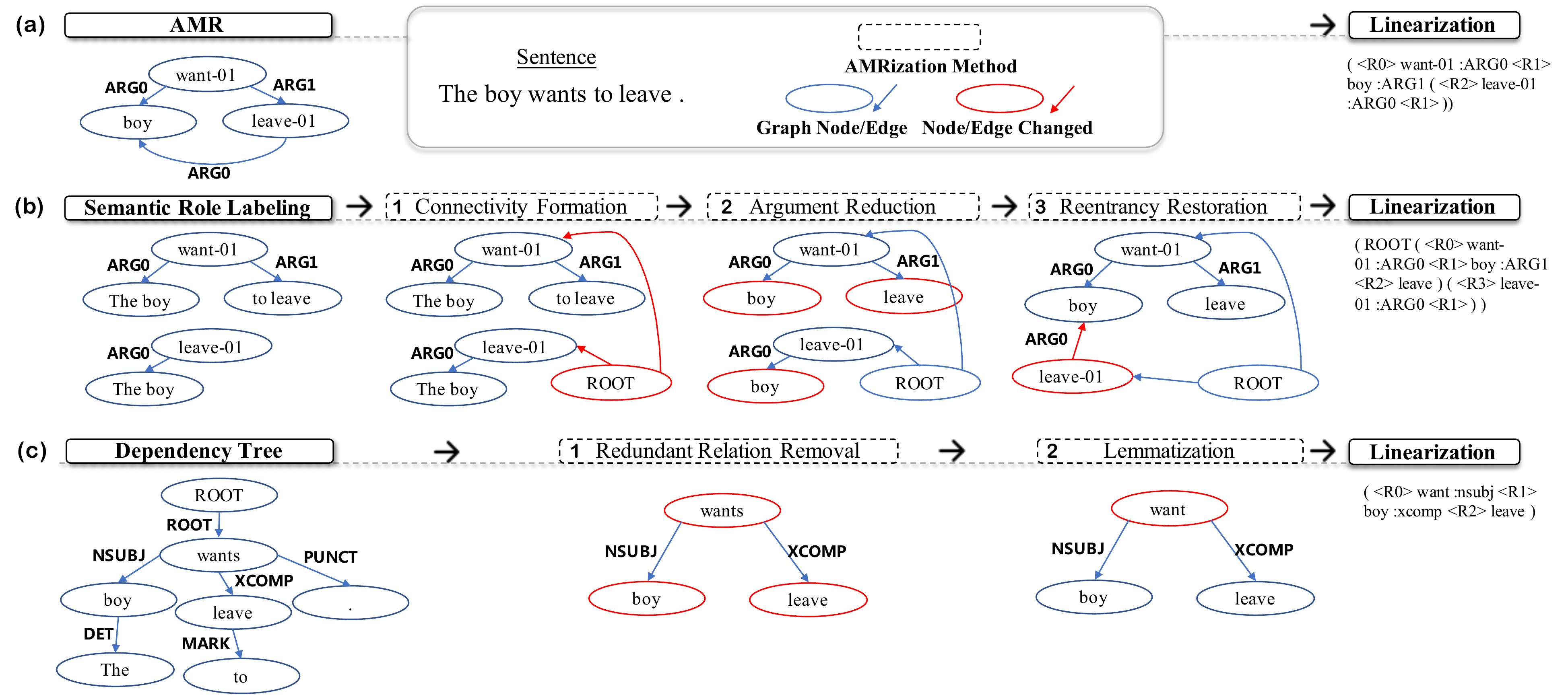}
    \caption{Illustration of AMRization methods and Graph Linearization. The source sentence is ``The boy wants to leave."}
    \label{fig:amrization}
\end{figure*}

As shown in Figure \ref{fig:procedure}, in this paper,  we propose a principled method to select auxiliary tasks (Section \ref{sec: task selection}), AMRize them into PseudoAMR (Section \ref{sec: amrization}) and train PseudoAMR and AMR effectively (Section \ref{sec: training paradigm}) to boost AMR parsing.
We formulate both PseudoAMR and AMR parsing as the sequence-to-sequence generation problem.
Given a sentence $x = [x_i]_{1 \leq i\leq N}$, the model aims to generate a linearized PseudoAMR or AMR graph $y = [y_i]_{1 \leq i\leq M}$ (the right part of Figure \ref{fig:amrization}) with a product of conditional probability:
\begin{equation*}
    P(y) = \prod_{i=1}^Mp(y_i|(y_1,y_2,...,y_{i-1}),x)    
\end{equation*}

\subsection{Auxiliary Task Selection}
\label{sec: task selection}
When introducing auxiliary tasks for AMR parsing, the selected tasks should be   formally or semantically related to AMR, thus the knowledge contained in them can be transferred to AMR parsing.
Based on this principle of relevance, we choose semantic role labeling (SRL) and dependency parsing (DP) as our auxiliary tasks. We involve Translation and Summarization tasks for comparison.

\paragraph{Semantic Role Labeling} 
SRL aims to recover the predicate-argument structure of a sentence, which can enhance AMR parsing, because:
(1) Recovering the predicate-argument structure is also a sub-task of AMR parsing. As illustrated in Figure \ref{fig:amrization}(a,b), both AMR and SRL locate the predicates (``want'', ``leave'') of the sentence and conduct word-sense disambiguation. Then they both capture the multiple arguments of center predicate. 
(2) SRL and AMR are known as shallow and deep semantic parsing, respectively.
It is reasonable to think that the shallow level of semantic knowledge in SRL is useful for deep semantic parsing.

\paragraph{Dependency Parsing} DP aims to parse a sentence into a tree structure, which represents the dependency relation among tokens. 
The knowledge of DP is useful for AMR parsing, since:
(1) Linguistically, DP (syntax parsing task) can be the precursor task of AMR (semantic parsing).
(2) The dependency relation of DP is also related to semantic relation of AMR, e.g., as illustrated in Figure \ref{fig:srl_dp_amr}(c), ``NSUBJ'' in DP usually represents ``:ARG0'' in AMR. Actually, they both correspond to the agent-patient relations in the sentence.
(3) DP is similar to AMR parsing from the perspective of edge prediction, because both of them need to capture the relation of nodes (tokens/concepts) in the sentence.

\subsection{AMRization}

\label{sec: amrization}
Although SRL and DP are highly related to AMR parsing, there still exists gaps between them,
e.g., SRL annotations may be disconnected, while AMR is always a connected graph.
To bridge these gaps, we transform them into PseudoAMR, which we call AMRization.

\subsubsection{Transform SRL to PseudoAMR}
We summarize typical gaps between SRL and AMR as:
(1) \textit{Connectivity}. AMR is a connected directed graph while the structure of SRL is a forest. 
(2) \textit{Span-Concept Gap}. Nodes in AMR graph represent concepts (e.g., ``boy'') while that of SRL are token spans (e.g., ``the boy'', ``that boy''). Actually all the mentioned token spans correspond to the same concept.
(3) \textit{Reentrancy}. Reentrancy is an important feature of AMR as shown in Figure~\ref{fig:amrization}(a), the instance boy is referenced twice as ARG0. The feature can be applied to conduct coreference resolution. However, there is no reentrancy in SRL.
To bridge such gaps, we propose \textbf{Connectivity Formation}, \textbf{Argument Reduction}   and \textbf{Reentrancy Restoration} to transform SRL into PseudoAMR. 

\paragraph{Connectivity Formation}
\label{sec:graph formalization} 
To address the connectivity gap, we need to merge all SRL trees into a connective graph. Note that the merging doesn't guarantee correctness in semantic level. As shown in Figure~\ref{fig:amrization}(b-1), we first add a virtual root node, then generating a directed edge from the virtual root to each root of SRL trees, thus the SRL annotation becomes a connected graph.

\paragraph{Argument Reduction}
To address the Span-Concept Gap, 
as shown in Figure~\ref{fig:amrization}(b-2), if the argument of current predicate is a span with more than one token, we will replace this span with its head token in its dependency structure. Thus token spans ``the boy'', ``that boy'' will be transformed to ``boy'', more similar to the corresponding concept. Similar method has been to applied by \cite{zhang-etal-2021-comparing} to find the head of token spans of argument.

\paragraph{Reentrancy Restoration} 
For the reentrancy gap, we design a heuristic algorithm based on DFS to restore reentrancy in SRL. 
As shown in Figure~\ref{fig:amrization}(b-3), the core idea of the restoration is that we create a variable when the algorithm first sees a node. If the DFS procedure meets node with the same name, the destination of current edge will be redirected to the variable we have created at first. Please refer to Appendix~\ref{alg:reen} for the pseudo code of the reentrancy restoration. 

\begin{figure}[t]
    \centering
    \includegraphics[width=1\linewidth]{ 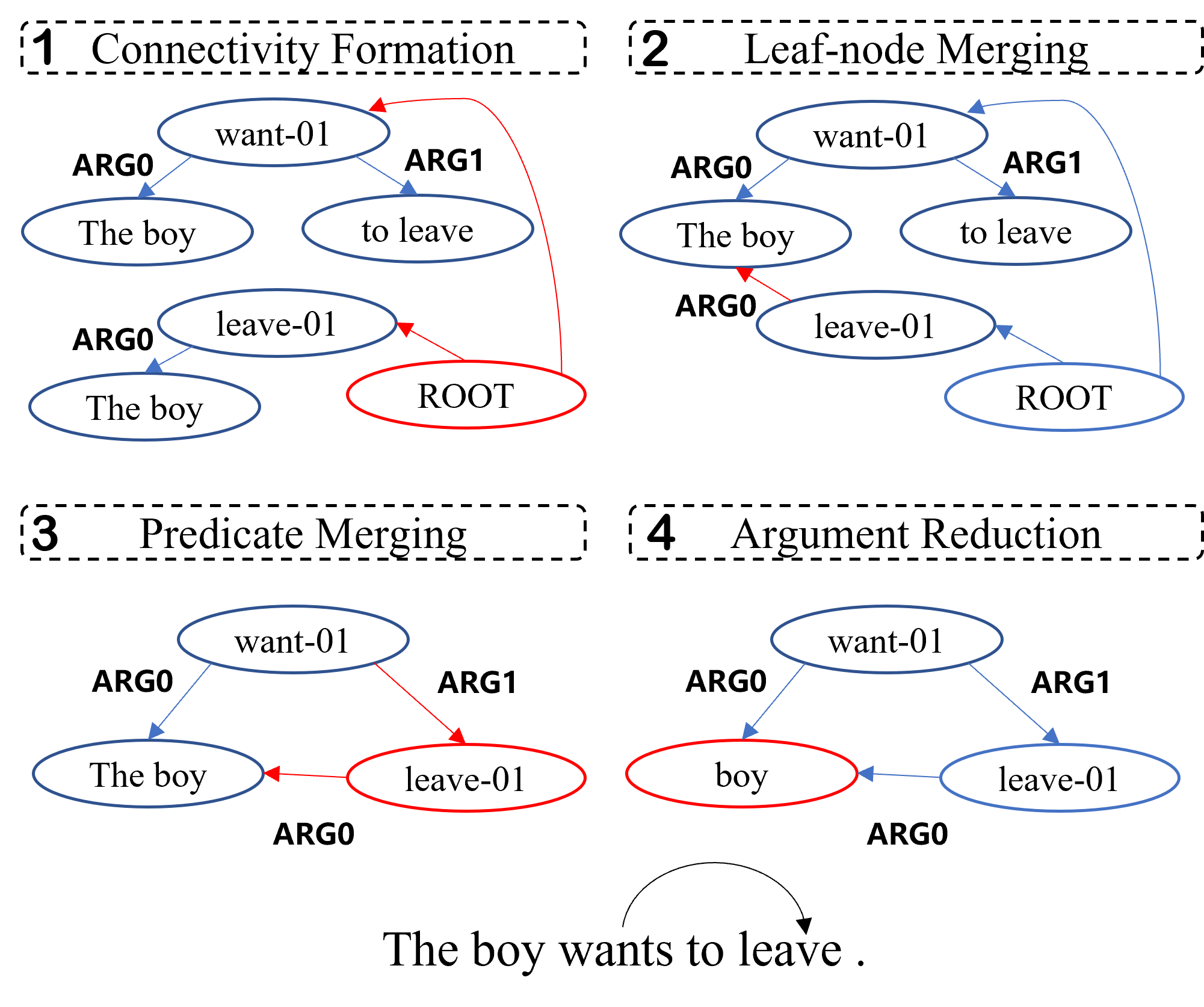}
    \caption{Illustration of Dependency Guided Restoration. In step 2, leaf-nodes ``The boy'' are merged. In step 3, none-leaf node ``leave-01'' is merged with leaf-node ``to leave'' since ``leave-01'' appears in word span ``to leave'' and  word ``leave'' depends on word ``want''.}
    \label{fig:dpg}
\end{figure}

\paragraph{Dependency Guided Restoration}
The previous restoration algorithm can not guarantee the merging of nodes agrees to the meaning of reentrancy in AMR since it merges concept according to their appearance order in the SRL structure. And it does not handle the merging of predicates. As shown in Figure~\ref{fig:amrization}(b-3), the node ``leave'' and ``leave-01'' should be merged, however we can't get this information directly from the SRL annotations. We therefore propose another restoration method based on the dependency structure of the corresponding sentence of the SRL as illustrated in Figure~\ref{fig:dpg}

This restoration algorithm takes the result of previous Connectivity Formation as input. It first merges the leaf-nodes corresponding to the same token. This step is accurate since leaf-nodes' merging will not bring divergence. The second step is to merge predicate nodes. For all sub-trees of the root node, it first check whether one predicate appear in others' argument span and whether the predicate directly depend on the span's predicate. If both two conditions are satisfied, the algorithm will merge the predicate and the span to one node. Last, it removes the root node and root-edges if the graph remains connected after removing. 

\subsubsection{Transform Dependency Structure to PseudoAMR} 
We summarize the gaps between Dependency Tree and AMR as:
(1) \textit{Redundant Relation}. Some relations in dependency parsing focus on syntax, e.g., 
``:PUNCT'' and ``:DET'', which are far from semantic relations in AMR. 
(2) \textit{Token-Concept Gap}. The basic element of dependency structure is token while that of AMR is the concept, which captures deeper syntax-independent semantics. 
We use \textbf{Redundant Relation Removal} and \textbf{Token Lemmatization} to transform the dependency structure to PseudoAMR to handle the gaps.

\paragraph{Redundant Relation Removal} 
For the Redundant Relation Gap, we remove some relations which are far from the sentence's semantics most of the time, such as ``PUNCT'' and ``DET''. As illustrated in Figure \ref{fig:amrization}(c-1), by removing some relations of the dependence, the parsing result become more compact compared with original DP tree, forcing the model to ignore some semantics-unrelated tokens during seq2seq training.

\paragraph{Token Lemmatization} 
As shown in Figure \ref{fig:amrization}(c-2),
for Token-Concept Gap, we conduct lemmatization on the node of dependency tree based on the observation that the affixes of single word do not affect the concept it corresponds to.  Together with the smart-initialization \citep{bevil-spring} by setting the concept token's embedding as the average of the subword constituents, the embedding vector of lemmatized token (`want') becomes closer to the vector concept (`want-01') in the embedding matrix, therefore requiring the model to capture deeper semantic when conducting DP task.

\subsubsection{Linearization}
After all AMRization steps, the graph structure of SRL/DP also should be linearized before doing seq2seq training.
As depicted in the right part of Figure~\ref{fig:amrization},
we linearize the graph by the DFS-based travel, and use special tokens \textit{<R0>, ..., <Rk>} to indicate variables, and parentheses to mark the depth, which is the best AMR linearization method of \citet{bevil-spring}.

\subsection{Training Paradigm Selection}
\label{sec: training paradigm}
After task selection and AMRization, we still need to choose an appropriate training paradigm to train PseudoAMR and AMR effectively. We explore three training paradigms as follows:
\paragraph{Multitask training}
Following \citet{xu-seqpretrain,Damonte2021OneSP}, we use classic schema in sequence-to-sequence multitask training by adding special task tag at the beginning of input sentence and training all tasks simultaneously. The validation of best model is conducted only on the AMR parsing sub-task.
\paragraph{Intermediate training}
Similar to \citet{kun2020intermediate}, we first fine-tune the pretrained model on the intermediate task (PseudoAMR parsing), followed by fine-tuning on the target AMR parsing task under same training setting. 
\paragraph{Multitask \& Intermediate training} We apply a joint paradigm to further explore how different paradigms affect AMR parsing. We first conduct multitask training, followed by fine-tuning on AMR parsing. Under this circumstance, Multitask training plays the role as the intermediate task.

\begin{table*}[!t]
    \centering
    \resizebox{0.9\textwidth}{!}{%
    \begin{tabular}{llccccccccccc}
        \toprule
                ~& \multirow{2}{*}{Model} &\multirow{2}{*}{Extra Data} & \multirow{2}{*}{\textsc{smatch}} &
                \multirow{2}{*}{NoWSD} &
                \multirow{2}{*}{Wiki} &
                \multicolumn{3}{c}{Concept-related} & \multicolumn{3}{c}{Topology-related} \\
                \cmidrule(r){7-9} \cmidrule(r){10-12} 
        ~ & ~ & ~ & ~& ~ &~& Conc. & NER & Neg. & Unll. & Reen. & SRL \\
        \midrule

        \multirow{13}{*}{\rotatebox[origin=c]{90}{{AMR 2.0}}}    
        
        ~& \citet{cai2020amr}      &  N  & 78.7  & 79.2 &81.3& 88.1 &87.1&66.1&81.5&63.8&74.5 \\
        ~& \citet{astudillo2020transition} & N & 80.2  & 80.7 &78.8 & 88.1 & 87.5 & 64.5 &  84.2 & 70.3 & 78.2 \\
        ~& \citet{zhou2021amr}   &  70k        & 81.8  & 82.3 &78.8 &  88.7 & 88.5 & 69.7 & 85.5 & 71.1 & 80.8 \\
                                          
        ~& SPRING \citep{bevil-spring}       &N     & 83.8  & 84.4& \textbf{84.3} & 90.2 & 90.6 & 74.4  & 86.1 & 70.8 & 79.6 \\

         ~&SPRING (w/ silver) \citep{bevil-spring}    &200k     & 84.3 & 84.8& 83.1 & \textbf{90.8} & 90.5 & 73.6  &  86.7 &  72.4 & 80.5 \\
         
        ~&SPRING (Ours)     &N     & 84.0 & 84.3& 83.5 & 89.9 & 91.8 & 75.1  &  87.1 & 71.3 & 81.3 \\

        ~& ATP (w/ DP) &40k  & 85.0 & 85.4 &  84.1& 90.4 &92.5  & 74.7   &88.2  & 74.7 & 83.1 \\
        
        ~& ATP (w/ SRL) &40k &85.1   &\textbf{85.6} &  83.6 &90.4  & 91.4  & \textbf{75.7}  & 88.2 & \textbf{75.0}  &  \textbf{83.5}   \\
        
        ~& ATP (w/ SRL$^D$) &40k & \textbf{85.2}&\textbf{85.6}&84.2&90.7&\textbf{93.1}&74.9&\textbf{88.3}&74.7&83.3 \\

        \cmidrule(r){2-12}

        
        ~& Graphene 4S$^E$  \citep{lam2021ensembling}   &200k  &84.8   & 85.3 &  83.9& 90.6 & 92.2  & \textbf{75.2}   &  88.0  & 71.4 & 83.5   \\

        ~& Structure-aware$^E$ \citep{saft} &47k &84.9 &-&-&-&-&-&-&-&-\\

        ~& ATP (w/ SRL) $^E$   &40k  &\textbf{85.3}   & \textbf{85.7} &  83.9& 90.7 & 92.2  & 75.0   &  \textbf{88.4}  & 75.0 & \textbf{83.6}   \\
        
        ~& ATP (w/ SRL$^D$) $^E$   &40k  &\textbf{85.3}   & \textbf{85.7} &  \textbf{84.0}& \textbf{90.8} & \textbf{92.7}  & 74.7   &  \textbf{88.4}  & \textbf{75.1} & \textbf{83.6}   \\

        \midrule
        \multirow{7}{*}{\rotatebox[origin=c]{90}{{AMR 3.0}}}
        ~& \citet{bevil-spring} (w/ silver) & 200k & 83.0 & 83.5& \textbf{82.7} & \textbf{89.8} & 87.2 & 73.0  & 85.4 & 70.4 & 78.9 \\
        ~& ATP (w/ DP) & 40k &\textbf{83.9}&\textbf{84.3}&81.6&89.7&\textbf{89.2}&73.0&\textbf{87.0}&73.7&82.3  \\
         ~& ATP (w/ SRL) &40k &\textbf{83.9} &\textbf{84.3}&81.0 &89.7 &88.4 &\textbf{73.9}  &\textbf{87.0}&\textbf{73.9}&\textbf{82.5} \\

        \cmidrule{2-12}
        
        ~& Graphene 4S$^E$  \citep{lam2021ensembling}   &200k  &83.8 & 84.2& \textbf{81.9} & \textbf{90.1} & 88.3 & \textbf{74.6}  & 86.9 & 70.2 & 82.5   \\

        ~& Structure-aware$^E$ \citep{saft} &47k &83.1&-&-&-&-&-&-&-&-\\
        
        ~& ATP (w/ SRL)$^E$ &40k & \textbf{84.0}&\textbf{84.5}&80.7&90.0&\textbf{88.9}&73.1&\textbf{87.1}&\textbf{73.9}&\textbf{82.6} \\

        \bottomrule
    \end{tabular}
    } 
    \caption{\textsc{smatch} and fine-grained F1 scores on AMR 2.0 and 3.0. $^D$ denotes model using Dependency Guided Restoration. $^E$ denotes result with model ensemble (the details of the ensembling models are described in Appendix \ref{app:baselines}). We conduct ensembling by averaging the models from three random seeds following \citet{saft}.}
    \label{tab:main_results2.0}
\end{table*}

\section{Experiments}
\subsection{Datasets}
\paragraph{AMR Datasets} We conducted out experiment on two AMR benchmark datasets, AMR 2.0 and AMR 3.0. AMR2.0 contains $36521$, $1368$ and $1371$ sentence-AMR pairs in training, validation and testing sets, respectively. AMR 3.0 has $55635$, $1722$ and $1898$ sentence-AMR pairs for training validation and testing set, respectively. We also conducted experiments in out-of-distribution datasets (BIO,TLP,News3) and low-resources setting.

\paragraph{Auxiliary Task Datasets} Apart from DP/SRL, we choose NLG tasks including summarization and translation to evaluate the contributions of auxiliary tasks. Description of datasets is listed Appendix~\ref{app:dataset_des}.

\subsection{Evaluation Metrics}
We use the Smatch scores \citep{cai-smatch}  and further the break down scores \cite{dam-smatch-incremental} to evaluate the performance.

To fully understand the aspects where auxiliary tasks improve AMR parsing, we divide the fine-grained scores to two categories: \textbf{1) Concept-Related} including Concept, NER and Negation scores, which care more about concept centered prediction. \textbf{2) Topology-Related} including Unlabeled, Reentrancy and SRL scores, which focus on edge and relation prediction. NoWSD and Wikification are listed as isolated scores because NoWSD is highly correlated with Smatch score and wikification relies on external entity linker system.

\subsection{Experiment Setups}
\paragraph{Model Setting}
We use current state-of-the-art Seq2Seq AMR Paring model SPRING \citep{bevil-spring} as our main baseline model and apply BART-Large \citep{lew-bart} as our pretrained model. Blink \citep{li-etal-2020-efficient} is used to add wiki tags to the predicted AMR graphs. We do not apply re-category methods and other post-processing methods are the same with \citet{bevil-spring} to restore AMR from token sequence. Please refer to Section \ref{training_details} from appendix for more training details.

\paragraph{AMRization Setting}
For SRL, we explore four AMRization settings. 1) Trivial. Concept :multi-sentence and relation :snt are used to represent the virtual root and its edges to each of the SRL trees. 2) With Argument Reduction. We use dependency parser from Stanford CoreNLP Toolkit \cite{corenlp} to do the argument reduction.  3) With Reentrancy Restoration 4) All techniques.

For DP, we apply four AMRization settings 1) Trivial. Extra relations in dependency tree are added to the vocabulary of BART 2) With Lemmatization. We use NLTK \citep{bird-2006-nltk} to conduct token lemmatization 3) With Redundant Relation Removal. We remove PUNCT, DET, MARK and ROOT relations. 4) All techniques.

\subsection{Main Results} 
We report the result (ITL + All AMRization Techniques) on benchmark AMR 2.0 and 3.0 in Table~\ref{tab:main_results2.0}. On AMR 2.0, our models with DP or SRL as intermediate task gains consistent improvement over the SPRING model by a large margin (1.2 Smatch) and reach new state-of-the-art for single model (85.2 Smatch). Compared with SPRING with 200k extra data, our models achieve higher performance with much less extra data (40k v.s. 200k), suggesting the effectiveness of our intermediate tasks. We also compare our models with contemporary work \citep{lam2021ensembling,saft}. It turns out that our ensemble model beats its counterpart with less extra data, reaching a higher performance (85.3 Smatch). In fact, even without ensembling, our model still performs better than those ensembling models and the model using Dependency Guided Restoration method achieves higher performance than the trivial one, showing the effectiveness of our methods.

On AMR 3.0, Our models consistently outperform other models under both single model (83.9 Smatch) and ensembling setting (84.0 Smatch). Same as AMR 2.0, our single model reaches higher Smatch score than those ensembling models, revealing the effectiveness of our proposed methods. 

\paragraph{Fine-grained Performance} To better analyse how the AMR parser benefits from the intermediate training and how different intermediate tasks affect the overall performance. We report the fine-grained score as shown in Table~\ref{tab:main_results2.0}. We can tell that by incorporating intermediate tasks, considerable increases on most sub-metrics, especially on the Topology-related terms, are observed. On both AMR 2.0 and 3.0 our single model with SRL as intermediate task achieves the highest score in Unlabeled, Reentrancy and SRL metrics, suggesting that SRL intermediate task improves our parser's capability in Coreference and SRL. 

DP leads to consistent improvement in topology-related metrics, which also derives better result on NER sub-task (92.5 on AMR 2.0, 89.2 on AMR 3.0). We suppose that the ":nn" relation which signifies multi-word name entities in dependency parsing helps the AMR parser recognize multi-word name entities.
Generally speaking, AMR parser gains large improvement in Topology-related sub-tasks and NER by incorporating our intermediate tasks in terms of the Smatch scores.

\subsection{Exploration in Auxiliary Task Selection} 

\begin{table}[t]
    \centering
    \resizebox{0.45\textwidth}{!}{
\begin{tabular}{lccccc}
        \toprule
             Model &Extra & \textsc{Smatch} & Conc. & Topo.  \\
        \midrule
         \textbf{Ours (w/ NLG) } \\
         - w/ DialogSum & 13k & 84.5 &85.5 &81.5\\
         - w/ CNNDM   & 40k & 84.4 &85.5 &81.7 \\
         - w/ CNNDM   & 80k & 84.2 &85.1 &81.4 \\
         - w/ EN-DE   & 40k &84.4 &85.3&81.5 \\ 
         - w/ EN-DE   & 80k &84.4& 85.4 &81.4 \\ 
         - w/ EN-DE   & 200k&84.2&84.6 &81.2 \\ 
         - w/ EN-DE   & 400k &83.6 &84.9 & 80.6  \\
         \textbf{Ours (w/ Parsing) } \\
         - w/ DP & 40k & 85.0&\textbf{85.9}&   82.0 &  \\    
         - w/ SRL & 40k  &\textbf{85.1} &85.8 &\textbf{82.2}   \\    
        \bottomrule
    \end{tabular}}
    \caption{Result of Task Selection. We first train BART on different auxiliary tasks for 10 epochs before AMR Parsing. We also report the average scores of Concept-related (Conc.) and Topology-related metrics (Topo.) }
    \label{tab:task_sel_small}
\end{table}

We explore how different tasks affect AMR parsing apart from DP and SRL. We involve two classic conditional NLG tasks, Summarization and Translation for comparison as shown in Table~\ref{tab:task_sel_small}. 

The comparison implies that SRL and DP are better auxiliary tasks for AMR Parsing even under the circumstance where their counterparts exploit far more data (40k v.s. 400k). In fact, the performance of MT drops while introducing more data, which contradicts with \citet{xu-seqpretrain} 's findings that more MT data can lead to better result when pretraining the \textit{raw Transformer model}. However, this is not surprising under the background of Intermediate-task Learning where we already have a pretrained model with large-scale pretraining. Whether the intermediate tasks' form fits for the target task is far more important than the amount of data in the intermediate-task as also revealed by \citet{poth2021intermediate}. According to their observation, tasks with the most data (QQP 363k, MNLI 392k) perform far worse ( -97.4\% relative performance degradation at most) on some target tasks compared with tasks having much smaller datasets (CommonsenseQA 9k, SciTail 23k) which on the contrary give a positive influence. 

In conclusion, our findings suggest that the selection of intermediate task is important and should be closely related to AMR parsing in form, otherwise it would even lead to a performance drop for AMR parsing.

\section{Analysis}


\begin{figure}[!t]
    \centering
    \includegraphics[width=0.8\linewidth]{ 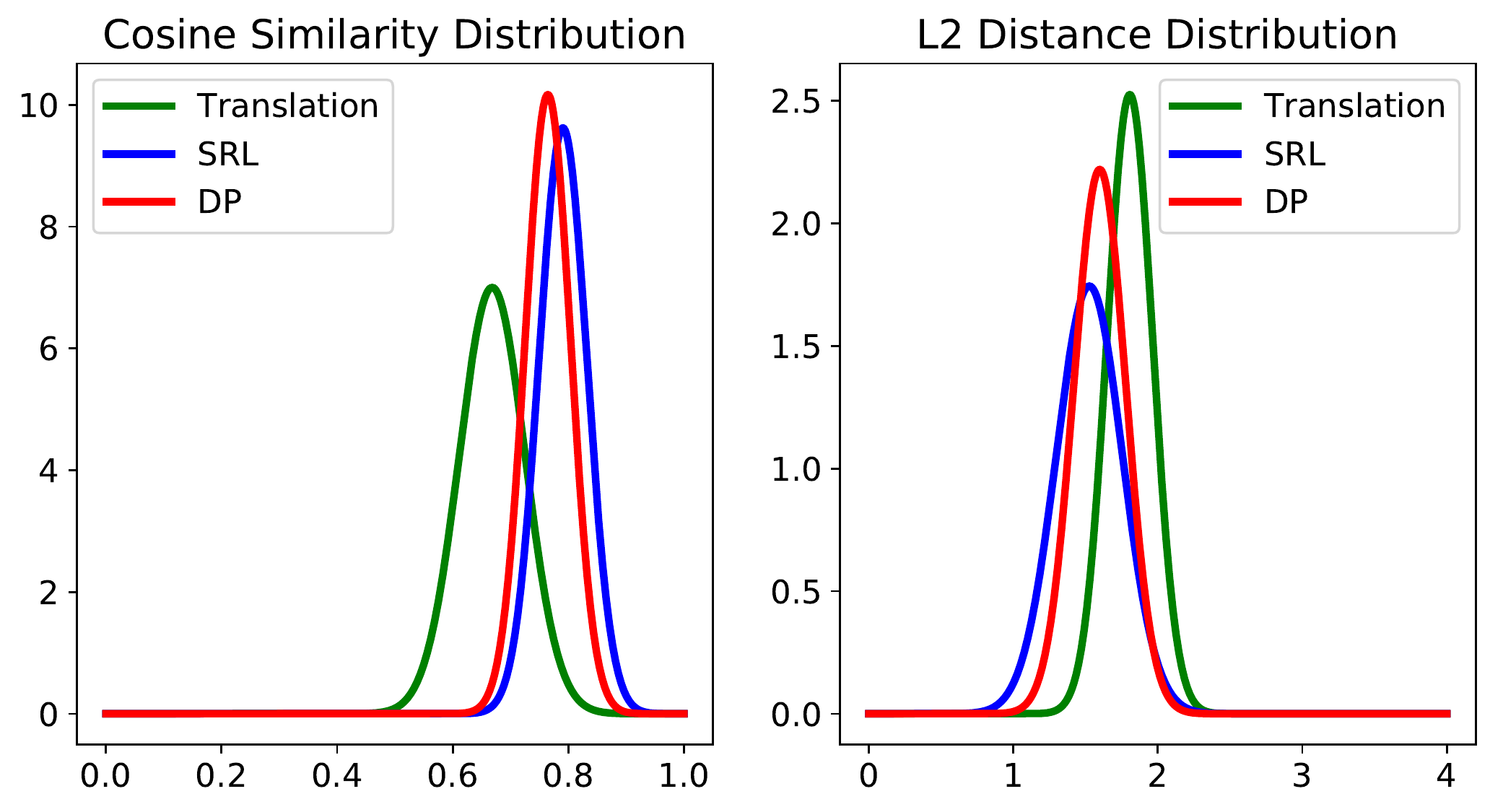}
    \caption{The distance distribution of sentences representation. SRL and DP consistently provide more similar sentence representation to AMR than Translation. The computation is illustrated in Figure~\ref{fig:comp} in appendix.  }
    \label{fig:repr}
\end{figure}

\renewcommand{\thefootnote}{\arabic{footnote}}

\subsection{More Similar Sentence Representation}
To examine how different auxiliary tasks affect AMR parsing, we collect the sentences' representation from different tasks' trained encoders\footnote{The computing process of sentences representation distance is illustrated in Figure~\ref{fig:comp} in appendix}. We use the average hidden state of the encoder's output as the sentence representation. We compute the Cosine Similarity and L2 distance between auxiliary tasks' representation and AMR's representation for same sentence. The test split of AMR 2.0 is used for evaluation. Finally, We apply Gaussian distribution to fit the distribution of distances and draw the probability distribution function curves as shown in Figure~\ref{fig:repr}. It turns out that under both distance metrics, SRL/DP consistently provide more similar sentence representation to AMR than Translation and SRL/DP are more similar to AMR parsing. It empirically justifies our hypothesis that semantically or formally related tasks can lead to a better initialization for AMR parsing.

\subsection{Ablation Study on AMRization Methods}

As shown in Table~\ref{tab:amrization_small}, we conduct ablation study on how different AMRization methods affect the performance AMR parsing. For both SRL and DP,  jointly adopting our AMRization techniques can further improve the performance of AMR parsing significantly, comparing to trivial linearization. The imperfect reentrancy restoration method leads to a significant improvement in terms of both the Topology and Concept related scores. It reveals that transformation of structure to mimic the feature of AMR can better the knowledge transfer between shallow and deep semantics.

As shown in Table~\ref{tab:amrization}, compared with jointly using the two techniques, it is worth noting that model with solely Reentrancy Restoration reaches highest fine-grained scores in especially on Reentrancy and SRL scores. To explore the reason why it surpasses adopting both techniques, we analyse the number of restored reentrancy. The result shows that about 10k more reentrancies are added when Argument Reduction (AR) is previously executed. It's expected since AR replaces the token span to the root token. Compared with token span, single token is more likely to be recognized as the correference variable according to the Reentrancy Restoration (RR) algorithm, thus generating more reentrancy, which might include bias to the model. This explains why solely using RR can lead to better results on SRL and Reen.

\begin{table}[t]
    \centering
    \resizebox{0.4\textwidth}{!}{
\begin{tabular}{lccc}
        \toprule
             Model & \textsc{Smatch} & Conc. & Topo. \\
        \midrule
         \textbf{Ours (w/ Semantic Role Labeling) } &84.5  & 85.5  & 81.6 \\
         - w/ Arg. Reduction(AR)   &  84.8  & 85.6  & 81.9\\
         - w/ Reen. Restoration(RR)   & 85.0  & 86.1  & \textbf{82.5} \\
         - w/ AR+RR &  85.1  & 85.8  & 82.2 \\ 
         - w/ AR+RR$^D$  &\textbf{85.2}  & \textbf{86.2}  & 82.1 \\ 
        \midrule
         \textbf{Ours (w/ Dependency Parsing) } & 84.4  & 84.7  & 81.7  \\
          - w/ Redundant Relation Removal (RRR)   &84.5  & 85.2  & 81.8   \\    
         - w/ Lemmatization (Lemma)     & 84.7  & 85.5  & 81.7\\    
         - w/ RRR + Lemma &\textbf{85.0}  & \textbf{85.9}  & \textbf{82.0}\\
        \bottomrule
    \end{tabular}}
    \caption{We report the average scores of Concept-related scores and Topology-related scores. The full scores are listed in Table~\ref{tab:amrization}. The improvement of involving all techniques against trivial linearization is significant with p < 0.005 for both SRL and DP.}
    \label{tab:amrization_small}
\end{table}

\begin{table}[t]
    \centering
\resizebox{0.4\textwidth}{!}{
\begin{tabular}{lccc}
        \toprule
          Model&Extra & \textsc{Smatch} \\
        \midrule
         
         \textbf{Ours (w/ Intermediate) } \\
         - w/ DP & 40k & 85.0 \\    
         - w/ SRL &   40k  &\textbf{85.1}     \\  
         - w/ DP,SRL & 80k & 84.7 \\  
         \textbf{Ours (w/ Multitask) } \\
         - w/ DP &40k & 83.7 \\
         - w/ SRL &40k &  83.6   \\
         - w/ DP,SRL &80k & 83.5   \\
         \textbf{Ours (w/ Multi. + Inter.) } \\
         - w/ DP &40k &84.1  \\
         - w/ SRL &40k & 84.1    \\
         - w/ DP,SRL &80k & 83.9    \\
        
        \bottomrule
    \end{tabular}}
    \caption{Analysis on Training Paradigms. Intermediate-task training is more suitable for AMR parsing than Multitask training}

    \label{tab:funetuning}
\end{table}

\begin{figure}[t]
    \centering
    \includegraphics[width=0.8\linewidth]{ 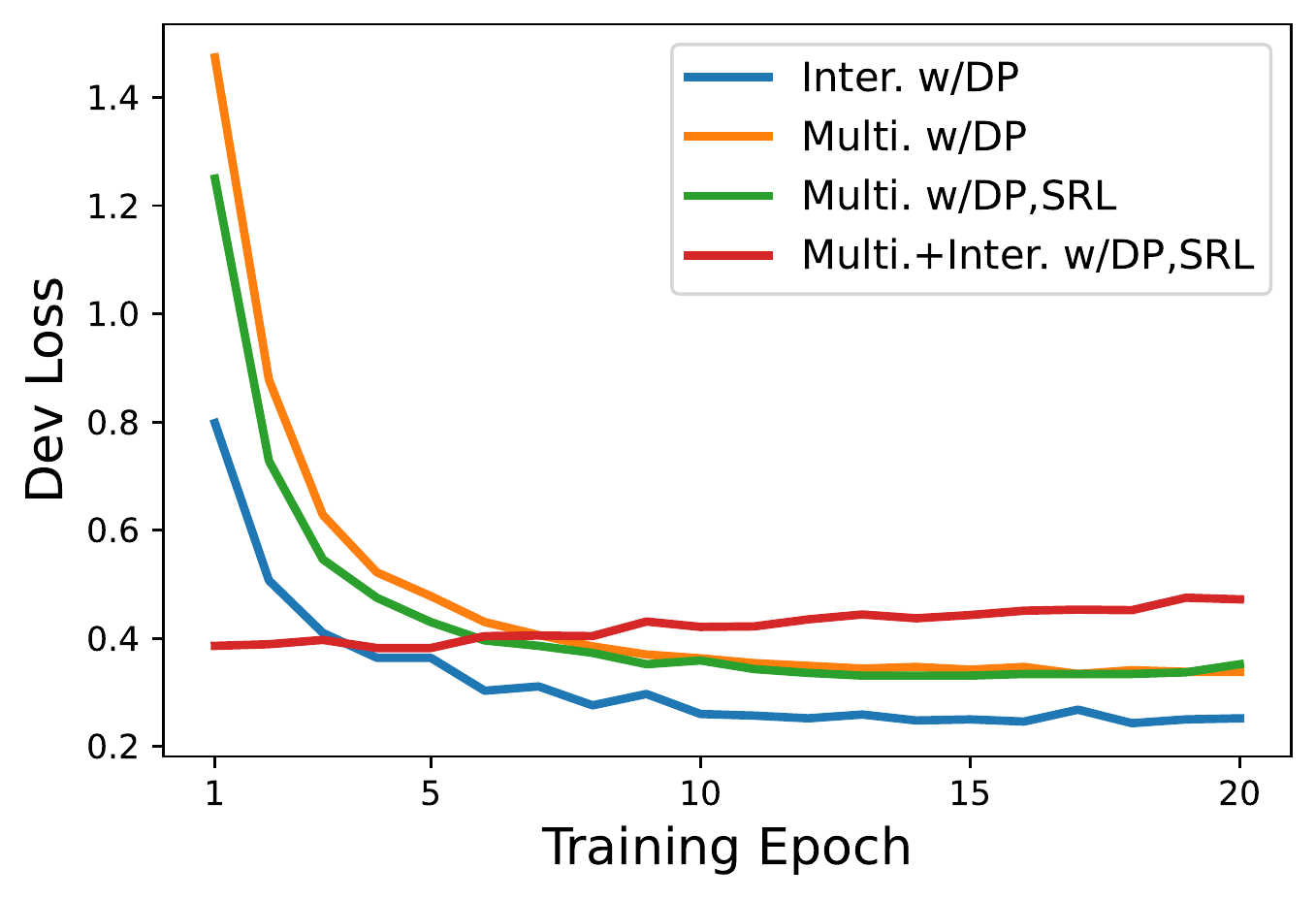}
    \caption{The loss curve on development set of AMR 2.0 for different training paradigms.}
    
    \label{fig:training-curve}
\end{figure}

\subsection{ITL Outweighs MTL}

We report the result of different fine-tuning paradigms in Table~\ref{tab:funetuning}. It justifies our assumption that classic multitask learning with task tag as previously applied in \citet{xu-seqpretrain,Damonte2021OneSP} does not compare with intermediate training paradigm for AMR Parsing task.

As shown in Figure~\ref{fig:training-curve}, Intermediate-task training provides a faster and better converging process than MTL. We assume this is due to the huge gap between AMR parsing and auxiliary tasks which may harm the optimization process of MTL. The process of optimizing all auxiliary tasks simultaneously may introduce noise to AMR Parsing. 

We also find that under the setting of ITL, sequentially training SRL and DP tasks did not bring further improvement to AMR parsing. We guess this is due to the catastrophic forgetting problem. Further regularization during training might help the model progressively learn from different auxiliary tasks and relieve catastrophic forgetting.

\subsection{Exploration in Out-of-Distribution Generalization} \begin{table}[t]
    \centering
    \footnotesize
\begin{tabular}{llccc}
        \toprule
          Model&BIO & TLP &News3 \\
        \midrule
         SPRING & 59.7 & 77.3 & 73.7\\
         SPRING + silver &59.5 & 77.5 &71.8 \\
         SPRING$^E$ & 60.5 & 77.9 &74.7 \\
         \textbf{Ours} &\textbf{61.2} & \textbf{78.9} & \textbf{75.4} \\
        
        \bottomrule
    \end{tabular}
    \caption{Analysis on OOD data. $^E$ denotes result given by the ensembling of models. Our model exploits SRL as the intermediate task.}

    \label{tab:ood}
\end{table}

Following \citet{bevil-spring,lam2021ensembling}, we assess the performance of our models when trained on out-of-distribution (OOD) data. The models trained solely on AMR 2.0 training data are used to evaluate out-of-distribution performance on the BIO, the TLP and the News3 dataset.

Table~\ref{tab:ood} shows the result of our out-of-distribution experiments. Our model surpass other models even the ensembled one\citep{lam2021ensembling}, creating new state-of-the-art for single model.

\subsection{Exploration in Low Resources Setting}
\begin{table}[t]
    \centering
\scalebox{0.8}{
\begin{tabular}{lllccc}
        \toprule
          ~&Model&BOLT  & LORELEI & DFA \\
        \midrule
        \multirow{2}{*}{\rotatebox[origin=c]{0}{Dev}}
         ~ &SPRING & 30.8  & 72.3 & 73.5\\
         ~ &\textbf{Ours} &\textbf{56.0}  & \textbf{73.9} & \textbf{76.1}\\
         \midrule
         \multirow{2}{*}{\rotatebox[origin=c]{0}{Test}}
         ~&SPRING &34.6  &73.8 & 71.1\\
         ~&\textbf{Ours} &\textbf{59.4}  & \textbf{74.5} &\textbf{74.3}\\
        \bottomrule
    \end{tabular}}
    \caption{Model Smatch scores in the low-resource setting. There are 1061, 4441, 6455 examples in the training set of BOLT, LORELEI and DFA, respectively. The model exploits SRL as the intermediate task.}

    \label{tab:few}
\end{table}

Since the annotation of AMR is both time and labor consuming, it raises our interests if we can improve the learning ability of AMR Parser under low resources setting. 

We set three low resources benchmarks  \textbf{BOLT, LORELEI, DFA} for AMR parsing based on the different sufficient degree of training examples. Detail of the datasets is described in Appendix~\ref{app:few-shots} . Compared with the AMR2.0 dataset which has 36521 training samples, the number of training samples in \textbf{BOLT, LORELEI, DFA} are 2.9\%, 12.2\% and 17.7\% of the number of AMR2.0. Table~\ref{tab:few} reports the result. Our model surpasses the SPRING model by a real large margin (about 25 Smatch) in the BOLT dataset which is the most insufficient in data and gains a consistent improvement on all datasets, suggesting that our pretraining method is effective under low resources conditions.

\section{Related Work}
\paragraph{AMR Parsing}
AMR parsing is a challenging semantic parsing task, since AMR is a deep semantic representation and consists of many separate annotations \cite{ban-AMR} (e.g., semantic relations, named entities, co-reference and so on).
There are four major methods to do AMR Parsing currently, sequence-to-sequence approaches \cite{ge-seq2seqamr, xu-seqpretrain, bevil-spring, HCL},
tree-based approaches \cite{zhang2019broad, zhang2019amr}, graph-based approaches \cite{lyu2018amr, cai2020amr} and transition-based approaches \cite{naseem2019rewarding, lee2020pushing, zhou2021amr}. 

There are two ways to incorporate other tasks to AMR Parsing. \citet{Goodman2016NoiseRA} builds AMR graph directly from dependency trees while \cite{ge-seq2seqamr} parse directly from linearized syntactic tree.  \citet{xu-seqpretrain} introduces Machine Translation, Constituency Parsing as pretraining tasks for Seq2Seq AMR parsing and \citet{Wu2021ImprovingAP} introduces Dependency Parsing for transition-based AMR parsing.  However all of them do not take care of the semantic and formal gap between the auxiliary tasks and AMR parsing.


\paragraph{Multitask \& Intermediate-task Learning}
Multi-task Learning (MTL) \cite{caruana1997multitask} aims to jointly train multiple related tasks to improve the performance of all tasks. Different from MTL, Intermediate-task Learning (ITL) is proposed to enhance pretrained models e.g. BERT by training on intermediate task before fine-tuning on the target task. Recent studies\citep{kun2020intermediate,poth2021intermediate} on ITL expose that choosing right intermediate tasks is important. Tasks that don't match might even bring negative effect to the target even if it has far more data. 

\citet{xu-seqpretrain,Damonte2021OneSP,procopio-etal-2021-sgl} utilize auxiliary tasks in a MTL fashion with specific task tags. \citet{bevil-spring,saft} adopt sliver training data in a ITL paradigm.  However, there is no work comparing ITL and MTL when introducing auxiliary tasks to enhance PTM-based AMR parser.


\section{Conclusion}
In this paper, We find that semantically or formally related tasks, e.g. SRL and DP are better auxiliary tasks for AMR parsing and can further improve the performance by proper AMRization methods to bridge the gap between tasks. And Intermediate-task Learning is more effective in introducing auxiliary tasks compared with Multitask Learning. Extensive experiments and analyses show the effectiveness and priority of our proposed methods. 

\section{Acknowledgements}
We thank all reviewers for their valuable advice. This paper is supported by the National Key Research and Development Program of China under Grant No.2020AAA0106700, the National Science Foundation of China under Grant No.61936012 and 61876004.

\section{Ethics Consideration}
We collect our data from public datasets that permit academic use and buy the license for the datasets that are not free. The open-source tools we use for training and evaluation are freely accessible online without copyright conflicts.

\newpage

\bibliography{acl2021}
\bibliographystyle{acl_natbib}

\clearpage
\appendix

\section{Algorithms}
\label{alg:reen}
\begin{algorithm}[htb]
\caption{Reentrancy Restoration for SRL} 
\hspace*{0.02in} {\bf Input:} 
Treenode:T\\
\hspace*{0.02in} {\bf Output:} 
Graph:G \\
\hspace*{0.02in} {\bf Description:} T is root node of the original SRL after node ROOT is added to form tree structure. G is the output graph with possible reentrancy restored.\\
\hspace*{0.02in} {\bf Global Variables:} Dict: V=\{\}. Here Dict is the official data structure of Python's dictionary.
\begin{algorithmic}[1]

\For{predicate in T.sons}
    \For{son in predicate.sons()}
        \If{son.name in V.keys()}
            \State son = V[son.name] 
            \State\# restore reentrancy 
\Else
\State V[son.name] = son
\EndIf
\EndFor
\EndFor
\State \Return T
\end{algorithmic}
\end{algorithm}

\section{Ensemble Models' Methods}
\label{app:baselines}

\paragraph{Graphene-4S$^E$} \citet{lam2021ensembling} make use of 4 SPRING models from different random seeds and their proposed graph ensemble algorithm to do the ensembling. They also include another ensemble model named Graphene All which includes four checkpoints from models of different architectures, SPRING\citep{bevil-spring}, APT\citep{zhou2021amr}, T5, and Cai\&Lam\citep{cai2020amr}. We do not report the score of Graphene All since it aggregates models with different inductive bias while our ensemble model only use models from one structure. It is out of the scope for fair comparison.

\paragraph{Structure-aware$^E$} \citet{saft} use ensemble results from 3 models' combination to generate the ensemble model.

\paragraph{Ours (w/ SRL)$^E$} We use the setting the same as \citet{saft}, we use the average of three models' parameters as the ensemble model.

\section{Auxiliary Datasets Description}
\label{app:dataset_des}
\subsection{Summarization}

\paragraph{\textsc{CNN/DM}\citep{Hermann2015TeachingMT}}The CNN / DailyMail Dataset is an English-language dataset containing news articles as written by journalists at CNN and the Daily Mail. The dataset is widely accepted as benchmark to test models' performance of summarizing .
\paragraph{\textsc{DialogSum}\citep{chen-etal-2021-dialogsum}}
The Real-Life Scenario Dialogue Summarization (\textsc{DialogSum}), is a large-scale summarization dataset for dialogues. Unlike \textsc{CNN/DM} which focuses on monologue news summarization, \textsc{DialogSum} covers a wide range of daily-life topics in the form of spoken dialogue. We use all the training data (13k) to conduct the intermediate training.

\subsection{Translation}

\paragraph{\textsc{WMT14 EN-DE}} We select the first 40k,80k,200k and 400k training examples from WMT14 EN-DE training set to form EN-DE translation intermediate tasks.

\subsection{Dependency Parsing}

\paragraph{\textsc{Penn Treebank}\citep{ptb3}} The Penn Treebank (PTB) project selected 2,499 stories from a three year Wall Street Journal (WSJ) collection of 98,732 stories for syntactic annotation. We only utilize the dependency structure annotations to form our intermediate dependency parsing task. There are 39,832 (\textasciitilde40k) sentences.

\subsection{Semantic Role Labeling}
\paragraph{\textsc{OntoNotes}\citep{Weischedel2017OntoNotesA}} The OntoNotes project is built on two resources, following the \textsc{Penn Treebank}\citep{ptb3} for syntax and the \textsc{Penn PropBank} for predicate-argument structure. We select 40k sentences with SRL annotations to form intermediate task.

\section{Low-resource Datasets Description}
\label{app:few-shots}
 We set three Low-resource Learning benchmark for AMR parsing:
 \begin{enumerate}
     \item  \textbf{BOLT} Using only the BOLT split of AMR data of AMR2.0 dataset. The training, validation and test data each has 1061, 133 and 133 amrs respectively.
     \item  \textbf{LORELEI} Using only the LORELEI split of AMR data of AMR3.0 dataset. The training,validation and test data each has 4441, 354 and 527 amrs respectively. 
     \item \textbf{DFA} Using only the DFA split of AMR data of AMR2.0 dataset. The training, validation and test data each has 6455, 210 and 229 amrs respectively.
 \end{enumerate}
 Compared with the AMR2.0 dataset which has 36521 training samples, the number of training samples in \textbf{BOLT, LORELEI, DFA} are 2.9\%, 12.2\% and 17.7\% of the number of AMR2.0.

\section{Training Details}
\label{training_details}
 We tune the hyper-parameters on the SPRING baseline, and then adding the auxiliary data using just those hyper-parameters without any changing.

We use RAdam \citep{liu-RAdam} as our optimizer, and the learning rate is $3e^{-5}$. Batch-size is set to 2048 tokens with 10 steps accumulation. The dropout rate is set to 0.3.

\begin{table}[h]
    \centering
    \footnotesize
    \resizebox{0.5\textwidth}{!}{
\begin{tabular}{ll}
        \toprule
          Parameter& Searching Space \\
        \midrule
         Learning rate & 1e-5, 3e-5, 5e-5, 1e-4 \\
         Batch-size & 256, 512, 1024, 2048, 4096 \\
         Grad. accu. & 10 \\
         Dropout & 0.1, 0.2, 0.3 \\
        \bottomrule
    \end{tabular}
    }
    \caption{Hyper-parameters searching space}

    \label{tab:ood}
\end{table}

 \begin{table*}[t]
    \centering
    \resizebox{0.9\textwidth}{!}{%
\begin{tabular}{llccccccccccc}
        \toprule
                ~& \multirow{2}{*}{Model} &\multirow{2}{*}{Extra Data} & \multirow{2}{*}{\textsc{smatch}} &
                \multirow{2}{*}{NoWSD} &
                \multirow{2}{*}{Wiki} &
                \multicolumn{3}{c}{Concept-related} & \multicolumn{3}{c}{Topology-related} \\
                \cmidrule(r){7-9} \cmidrule(r){10-12} 
        ~ & ~ & ~ & ~& ~ &~& Conc. & NER & Neg. & Unll. & Reen. & SRL \\
        \midrule

        \multirow{9}{*}{\rotatebox[origin=c]{90}{{AMR 2.0}}}     
         ~&SPRING (w/ silver) \citep{bevil-spring}    &200k     & 84.3 & 84.8& 83.1 & \textbf{90.8} & 90.5 & 73.6  &  86.7 &  72.4 & 80.5 \\

        ~& \textbf{Ours (w/ Semantic Role Labeling) } &40k &84.5 &84.9&84.0 &90.2 & 91.8&74.6&87.7 &74.2 &82.8 \\
        ~& - w/ Arg. Reduction(AR) &40k &84.8 &85.2&83.9 &90.4 &92.2 &74.2&88.1 & 74.5 &83.0  \\
        ~& - w/ Reen. Restoration(RR)  &40k &85.0 &85.4&83.5&90.6 & 92.1 &75.6&\textbf{88.2}&\textbf{75.5}&\textbf{83.7}\\
        ~& - w/ AR+RR  &40k    &\textbf{85.1}   &\textbf{85.6} &  83.6 &90.4  & 91.4  & \textbf{75.7}  & \textbf{88.2} & 75.0  &  83.5   \\
       
        ~& \textbf{Ours (w/ Dependency Parsing) } &40k  &84.4 &84.9& 82.9 &90.1&90.5 &73.5 &87.8 &74.3 &82.9 \\
        ~& - w/ Redundant Relation Removal (RRR) &40k & 84.5 & 	85.0	&83.5 	&90.2 	&91.2 	&74.3 	&88.0 &74.5 	&82.9  \\
        ~& - w/ Lemmatization (Lemma)  &40k & 84.7 	& 	85.2 &83.8	&90.2 	&91.2 	&75.0 	 	&88.0 &74.1 	&83.0   \\
        ~& - w/ RRR + Lemma  &40k  & 85.0          & 85.4 & \textbf{84.1} & 90.4& \textbf{92.5}  & 74.7  &  \textbf{88.2}  &  74.7 & 83.1 \\

        \bottomrule
    \end{tabular}
    }
    \caption{Full scores of ablation on AMRization methods.}

    \label{tab:amrization}
\end{table*}

\begin{figure*}[t]
    \centering
    \includegraphics[width=1\linewidth]{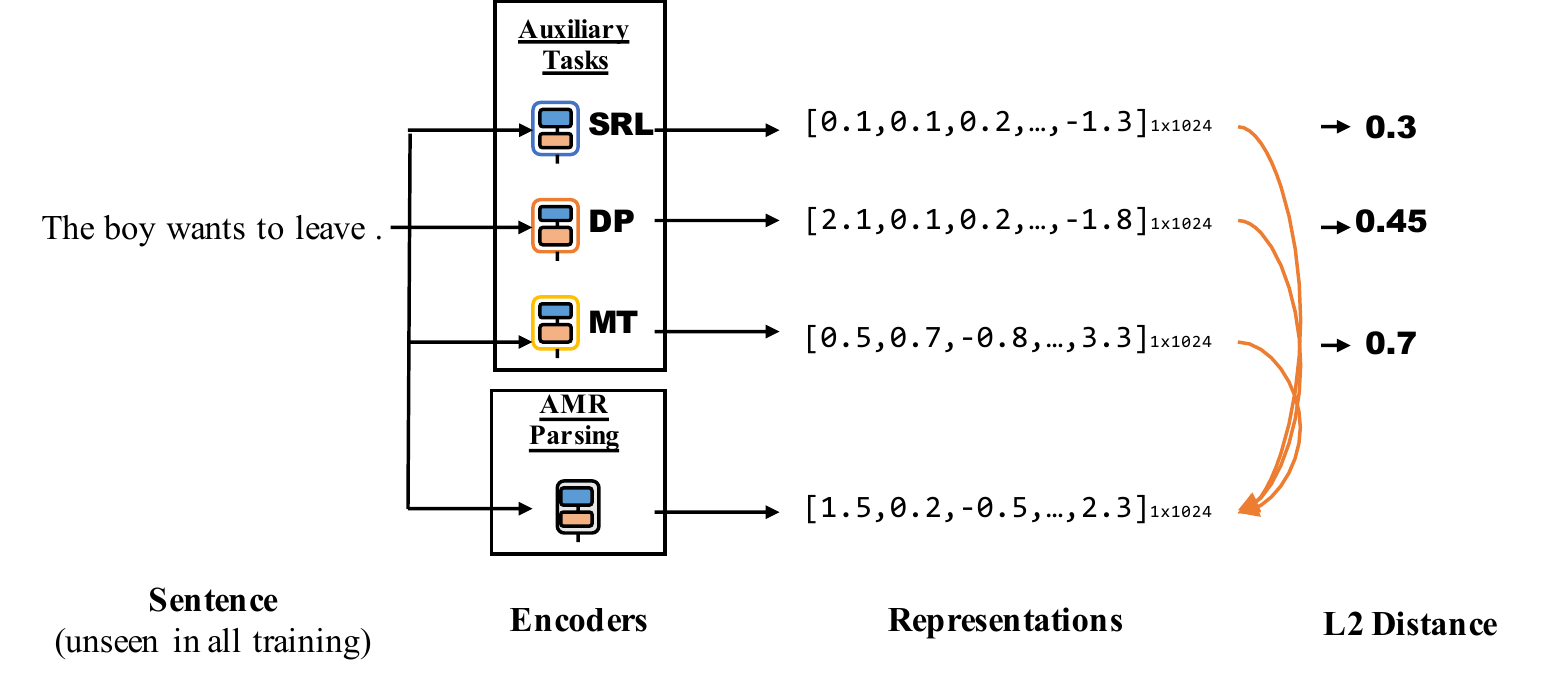}
    \caption{Illustration of how to compute sentence representation distance of different tasks. The sentences used for evaluate are never seen in the training of AMR Parsing and other auxiliary tasks. Cosine Similarity is computed the same way. We collect all sentences' distance of one encoder to draw the Gaussian distribution curve. }
    \label{fig:comp}
\end{figure*}

\end{document}